\documentclass{article}

\usepackage{microtype}
\usepackage{graphicx}
\usepackage{comment}
\usepackage{tabularx}
\usepackage{subfigure}
\usepackage{float}
\usepackage{lipsum}  
\usepackage{dsfont}
\usepackage{booktabs} 
\usepackage{multirow}
\usepackage{adjustbox}
\usepackage{hyperref}

\usepackage[accepted]{icml2024}

\usepackage{amsmath}
\usepackage{amssymb}
\usepackage{mathtools}
\usepackage{amsthm}

\usepackage[capitalize,noabbrev]{cleveref}

\theoremstyle{plain}

\theoremstyle{definition}

\theoremstyle{remark}

\usepackage[textsize=tiny]{todonotes}

\icmltitlerunning{Decoupling Feature Extraction and Classification Layers for Calibrated Neural Networks}

\begin{document}

\twocolumn[
\icmltitle{Decoupling Feature Extraction and Classification Layers for \\Calibrated Neural Networks}




\begin{icmlauthorlist}
\icmlauthor{Mikkel Jordahn}{DTU}
\icmlauthor{Pablo M. Olmos}{UC3M}
\end{icmlauthorlist}

\icmlaffiliation{DTU}{Cognitive Systems, Technical University of Denmark, Kongens Lyngby, Denmark}
\icmlaffiliation{UC3M}{Signal Processing Group (GTS), Universidad Carlos III de Madrid, Madrid, Spain}

\icmlcorrespondingauthor{Mikkel Jordahn}{mikkjo@dtu.dk}

\icmlkeywords{Machine Learning, ICML}

\vskip 0.3in
]



\printAffiliationsAndNotice{}  

\begin{abstract}
Deep Neural Networks (DNN) have shown great promise in many classification applications, yet are widely known to have poorly calibrated predictions when they are over-parametrized. Improving DNN calibration without comprising on model accuracy is of extreme importance and interest in safety critical applications such as in the health-care sector. In this work, we show that decoupling the training of feature extraction layers and classification layers in over-parametrized DNN architectures such as Wide Residual Networks (WRN) and Visual Transformers (ViT) significantly improves model calibration whilst retaining accuracy, and at a low training cost. In addition, we show that placing a Gaussian prior on the last hidden layer outputs of a DNN, and training the model variationally in the classification training stage, even further improves calibration. We illustrate these methods improve calibration across ViT and WRN architectures for several image classification benchmark datasets. 
\end{abstract}

\color{black}
\section{Introduction}
\label{submission}

\begin{figure*}[ht]
\vskip 0.2in
\begin{center}
\centerline{\includegraphics[width=\textwidth]{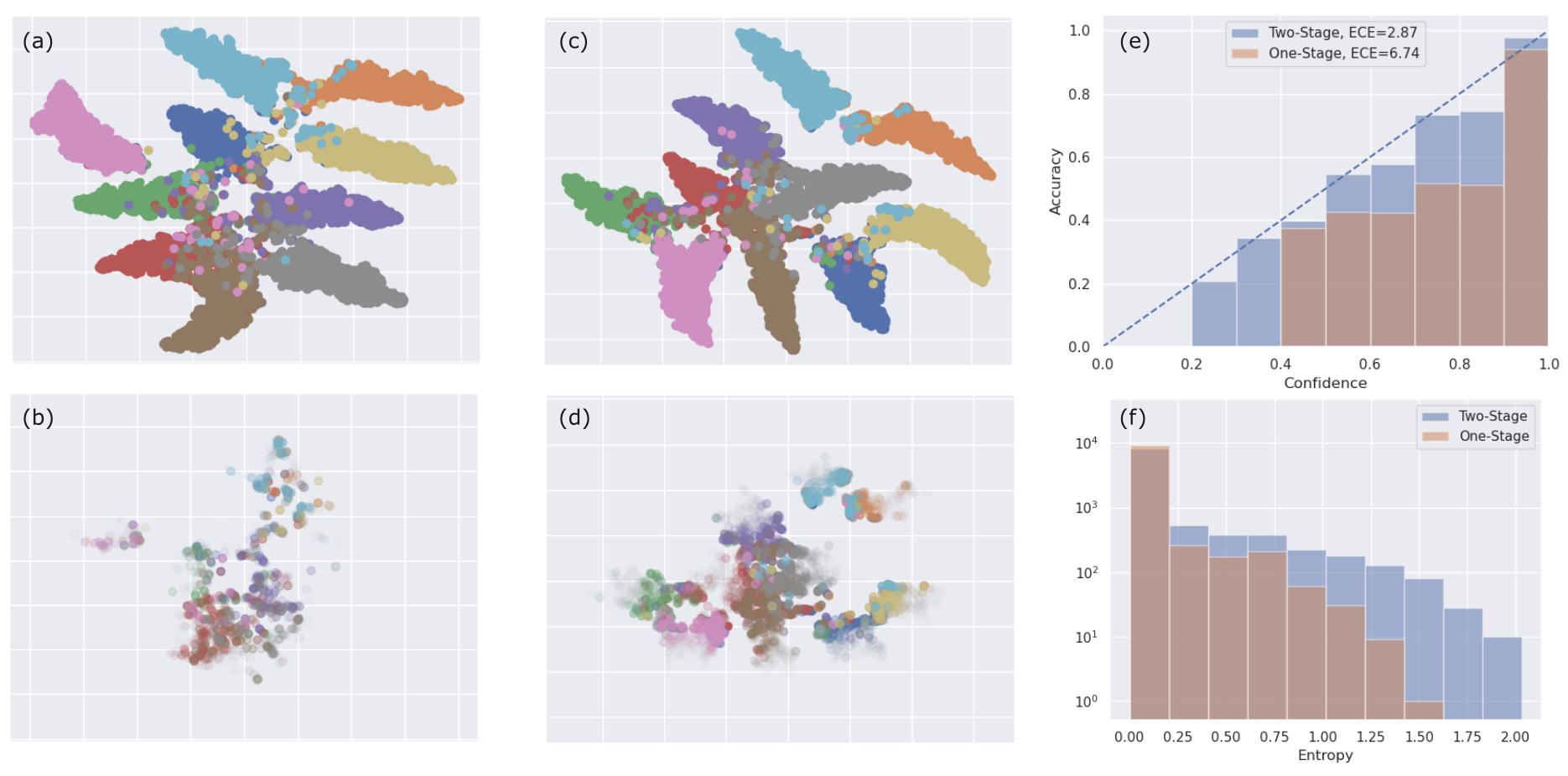}}
\caption{Last hidden layer outputs 2D t-SNE plots of test points from a WRN trained in \textbf{one stage} (left column), and \textbf{two stages} (center column) on CIFAR10 with accuracy $91.41\%$ and $91.61\%$ respectively. In the top row of t-SNE plots, (a) and (c), we plot all test points with equal color intensity, whilst in the second row, (b) and (d), higher color intensity indicates higher entropy of the prediction for a given point. Colors indicate which class the point truly belongs to. In (e) we show a calibration plot for the two models, and in (f) we show a histogram over the entropies of the predictions made by each model on the test set.}
\label{fig:intro}
\end{center}
\vskip -0.2in
\end{figure*}

Classification using machine learning is widely applied in many industries, ranging from application in medical diagnostics \cite{breast_cancer, healthcare} to being a crucial part of self-driving car systems \cite{lane_keeping, self_driving}. In applications with a safety or ethical aspect, it is not only the accuracy of the applied model that matters but also the calibration of the model. To give an example, if a machine learning model is applied in medical diagnostics, we simply do not just want to rely on the model's prediction, but rather the probability of this prediction. With calibration models, it is possible to refer to a doctor when model prediction uncertainty is high, which is not the case for poorly calibrated models. \\

Throughout this work, we refer to calibration in terms of the probabilities assigned to a prediction, i.e., if a perfectly calibrated model assigns probability $p$ of class $C$ in $A$ events, we would expect $p$ of those $A$ events to be class $C$. In the case of deep neural networks (DNN) for classification, it is well known that over-parameterized models under cross-entropy training tend to be overconfident \cite{TempScaling}. \\

Several methods have been proposed to create well-calibrated DNNs. Some of these methods are particular model specifications such as Bayesian Neural Networks (BNNs) \cite{Laplace_McKay, Dropout_BNN, Deep_Ensembles,Blei19,  SWAG}. Other methods have been proposed to \textit{re-calibrate} models post-hoc, i.e., after the model has been regularly trained, promising to retain model accuracy \cite{TempScaling, BTS}. A class of methods also uses data augmentation to regularize, thus creating calibrated models directly during training \cite{LabelSmoothing, MixUp}.

In this work, we show empirical evidence that the common practice of jointly training the feature extraction layers (convolution or attention layers) and the classification layers (fully connected layers) can lead to uncalibrated models. We then demonstrate that two-stage training procedures that decouple the training of the feature extraction and classification layers remarkably improve calibration. Finally, two training strategies are presented and analyzed: Two-Stage Training (TST) and Variational Two-Stage Training (V-TST). 

In TST, we first train a neural network end-to-end with cross-entropy loss (CE) until convergence.
In the second training stage, we freeze the feature extraction layers (convolution layers or attention layers) of the network, reinitialize the classifying fully connected (FC) layers, and re-train them on the same dataset. In V-TST, we additionally regularize the hidden feature space at the input of the last layer of the classifier with a Gaussian prior distribution and train the network with the evidence lower-bound (ELBO), enforcing better structure in the feature space and improving even further the model's calibration. 

We demonstrate how these methods significantly improve calibration metrics for on CIFAR10, CIFAR100 and SVHN and for different model architectures, in particular Wide Residual Networks (WRN) \cite{WRN} and Visual Transformers (ViT) \cite{ViT}.

In summary, our contributions are:
\begin{enumerate}
    \item We show that calibration in DNNs can be significantly improved if feature extraction layers are not jointly learnt with classification layers.
    \item We propose a two-stage training method that improves several calibration test metrics with respect to the same model trained in the usual end-to-end fashion.
    \item We show that placing a probabilistic prior on the final hidden layer outputs and variationally training the classification layers further improves model calibration.
\end{enumerate}

\section{A Motivational Example}

 Before presenting TST and V-TST, we provide a brief proof-of-concept experiment. In \cref{fig:intro} we demonstrate the effect of decoupling feature extraction training from classification layer training for a WRN architecture. In (a) we show the 2D t-SNE \cite{t-SNE} plot of the last hidden layer output $\mathbf{z}$ when the WRN is trained end-to-end using cross-entropy. In (b), the color intensity of the points depends on the classifier entropy $H(p(y|\mathbf{x}))$, so that zero entropy corresponds to zero color intensity (point not displayed). While the hidden space $\mathbf{z}$ is indeed structured by class (the overall classifier accuracy is $91.41\%$), this structure is mostly lost for those points with the largest uncertainty. For a calibrated classifier, we would expect that most points with the largest entropy correspond to cluster boundaries.

In (c) and (d), we show the same results for the WRN after re-training the classification FC layers from scratch in a second stage whilst the convolution layers are frozen. The overall test accuracy remains roughly the same ($91.61\%$), but (d) suggests a clearer separation of high uncertainty points (and a larger presence of them). The points with higher entropy are generally located in areas of the feature space where the class clusters are close to each other, a behaviour that suggests a better calibrated model. This is corroborated by the calibration plots \cite{TempScaling} in (e). From the entropy histograms in (f), we can observe that simply re-training the classifier has re-calibrated its confidence, decreasing the fraction of near-zero entropy points. 

\section{Two-Stage Training}

Based on these observations, we present two variations of our method for training calibrated DNNs: Two-Stage Training (TST) and Variational Two-Stage Training (V-TST). We show the algorithm for TST in \cref{alg:TST}.

\begin{algorithm}[H]
\centering
\caption{TST}
\begin{algorithmic}[1]
\STATE Init. DNN $M$ w. parameters $\{\beta, \phi\}$.
\STATE \textbf{Stage 1}: Train $M$ with CE loss on $D_{\text{train}}$ until convergence or early stopped.
\STATE Freeze parameters $\beta$ of $M$.
\STATE Re-init, FC layers of $M$ w. parameters $\{\theta, \nu\}$.
\STATE \textbf{Stage 2}: Train $\{\theta, \nu\}$ of $M$ with CE loss on $D_{\text{train}}$ until convergence.
\end{algorithmic}
\label{alg:TST}
\end{algorithm}

\begin{figure}[h]
\begin{center}
\centerline{\includegraphics[width=\columnwidth]{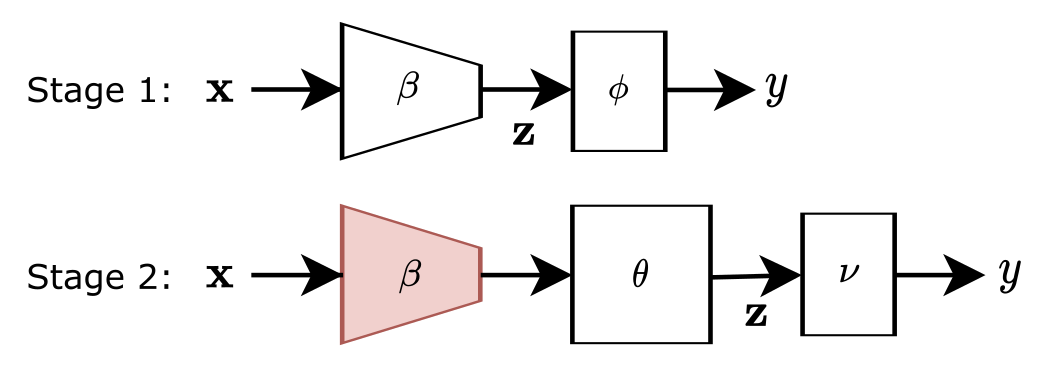}}
\caption{Block Diagram of Model $M$ and belonging parameters. Red background denotes frozen parameters.}
\label{fig:model_diagram}
\end{center}
\vskip -0.2in
\end{figure}

Here $\beta$ denotes the parameters of the feature extraction layers of DNN model $M$, $\phi$ are the initial FC layer parameters. For convenience to later describe V-TST, the parameters of the FC layers in Stage 2 are split into $\theta$ (all FC layers except the final logit layer) and $\nu$ the parameters of the logit layer itself. Note that the training data in Stage 1 and 2 are identical; therefore, our method is not a variation of pre-training and fine-tuning procedures. We show a block diagram of the model and parameters in \cref{fig:model_diagram} for clarification.

The classification layers trained in Stage 1 and 2 can have the exact same structure, as we consider in the experiment in \cref{fig:intro}. However, our experimental results suggest that even larger calibration improvements can be obtained by cross-validating the structure of the FC classification layers in Stage 2. In particular, appropriately choosing the dimensionality of the last hidden layer output $\mathbf{z}$ (the output of the last layer parametrized by $\theta$) can have a noticeable impact. In this regard, we hypothesize that further imposing structure on the feature space in the last hidden layer can be beneficial from the confidence calibration point of view. To verify such a hypothesis, we propose variational TST (V-TST).

\begin{figure}[H]
\centering
\begin{tikzpicture}[node distance={15mm}, thick, main/.style = {draw, circle}] 
\node[main] (Z) {$\mathbf{z}$}; 
\node[main] (Y) [below left of=Z] {$y$}; 
\node[main] (X) [below right of=Z] {$\mathbf{x}$}; 
\draw[->] (Z) -- (Y);
\draw[->, dashed] (X) to[out=90,in=0] (Z);
\end{tikzpicture} 
\caption{Latent Variable Model}
\label{fig:graph}
\end{figure}
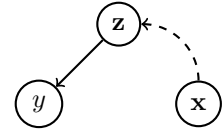

In V-TST we place a probabilistic prior on $\mathbf{z}$, and the FC layers parametrized by $\theta$ and $\nu$ are trained using an ELBO-like loss function instead of CE loss (while the parameters $\beta$ are fixed as in TST). More specifically, we consider the probabilistic model illustrated in \cref{fig:graph}. We assume a Gaussian latent prior $\mathbf{z}\sim \mathcal{N}(\mathbf{0},\mathbf{I})$ and a label-reconstruction probabilistic model given by
\begin{equation}
    p(y|\mathbf{z}) = \text{Cat}(\pi_{\nu}(\mathbf{z}))
\end{equation}
where $\pi_{\nu}(\mathbf{z})$ is the logits layer in TST. To enforce larger regularization in the $\mathbf{z}$ space, the prior distribution for $\mathbf{z}$ is kept simple and is not dependent on the image $\mathbf{x}$.

Classification and training in the V-TST probabilistic model are attained using a conditional variational posterior that is dependent on the classifier input $\mathbf{x}$:
\begin{align}\label{eq:qzx}
q(\mathbf{z}|\mathbf{x}) = \mathcal{N}(\mu_{\beta, \theta}(\mathbf{x}), \sigma_{\beta, \theta}(\mathbf{x})),
\end{align}
where $\mu_{\beta, \theta}(\mathbf{x})$ and $\sigma_{\beta, \theta}(\mathbf{x})$ are NNs that include the (frozen) feature extraction layers parametrized by $\beta$ and the trainable FC layers parametrized by $\theta$.  Given the variational posterior in \cref{eq:qzx}, the FC layers parametrized by $\theta$ and $\nu$ are trained to maximize the following lower bound:

\begin{align}
\begin{split}
    \log p(\mathbf{y})&\geq \text{ELBO}_{\theta,\nu}  = \int q(\mathbf{z}|\mathbf{x})\text{log}\frac{p(y|\mathbf{z})p(\mathbf{z})}{q(\mathbf{z}|\mathbf{x})}d\mathbf{z} \\
     = &\mathop{\mathbb{E}_{q(\mathbf{z}|\mathbf{x})}}\text{log}[p(y|\mathbf{z})] - D_{KL}(q(\mathbf{z}|\mathbf{x})||p(\mathbf{z})) 
    \label{eq:ELBO}
\end{split}
\end{align}

The ELBO in \cref{eq:ELBO} is optimized for all $(\mathbf{x},y)$ training pairs by standard mini-batch optimization combined with Monte Carlo (MC) sampling and Gaussian re-parameterization for the expectation w.r.t. $q(\mathbf{z}|\mathbf{x})$ as done in \citet*{VAE}. At prediction time, we similarly estimate $p(y|\mathbf{x})\approx\mathbb{E}_{q(\mathbf{z}|\mathbf{x})}[p(y|\mathbf{z})]$ using MC sampling.

\section{Related Work}

There exists a wide range of literature concerning designing and training well-calibrated neural networks with good uncertainty quantification. Firstly, the BNN literature holds in common that a prior is placed on all or some parameters of the neural network, and a posterior over these parameters is approximated. Several BNN methods have seen much interest in recent years, including Deep Ensembles \cite{Deep_Ensembles}, Laplace Approximation for Neural Networks \cite{Laplace_McKay}, Mean-Field BNNs \cite{MF_BNN, Blei19}, Node-Based BNNs \cite{Node_Rank1, Node_Helsinki} and MC-Dropout \cite{Dropout_BNN}.\\

TST, on the other hand, is a training method modification that falls into the implicit regularisation methods proposed for improving uncertainty quantification. There exist other implicit regularisation methods such as the Focal Loss \cite{Focal_Lin, Focal_Calib}, yet TST is not an alternative to these, but rather complementary. Another popular regularisation type is data augmentations for improved uncertainty quantification, such as Mix-Up \cite{MixUp} and Label-Smoothing \cite{LabelSmoothing}. Finally, post-hoc calibration methods exist, aiming to calibrate the soft-max probabilities of an already-trained neural network, namely Temperature Scaling (TS) \cite{TempScaling}.

In the recently proposed Variational Classifier by \citet*{VariationalClassification}, the authors also propose using variational inference for classifiers. In the work, the authors assume a latent prior $\mathbf{z}$ dependent on $y$, and training requires the use of an auxiliary discriminator since the latent distribution $q(\mathbf{z}|y)$ is implicit. This is only accessible by sampling from $\mathbf{z}\sim q(\mathbf{z}|\mathbf{x})$ for class samples $\mathbf{x}\sim p(\mathbf{x}|y)$. This results in a mini-max optimization objective over the ELBO. Further, the models are always trained end-to-end. V-TST on the other hand, has much less complex inference. In fact, V-TST can be viewed as a discriminative classifier with a stochastic final layer that requires training using ELBO instead of CE. All the complex and heavy feature extraction layers are trained in a supervised fashion. The V-TST generative process models exclusively the marginal label distribution $p(y)$, not $p(y|\mathbf{x})$, resulting in a simple ELBO objective for which training convergence is very fast. 

\citet*{GMLikelihoodLoss} is another work that proposes imposing a Gaussian structure on the feature space of a classifier, but they use a likelihood regularizer to enforce this structure rather than the ELBO, nor do they train in two stages and do not report on calibration improvements.

\section{Experiments}

\begin{table*}[!ht]
\caption{Test Metrics on In-Distribution Data Test-sets. For V-TST, $m=1$ and $m=10$ indicates using $1$ and $10$ samples in the MC approximation, respectively. Temp. WRN is the temperature scaled WRN. Bolded fonts indicate best performing model for given dataset.}
\label{tab:best_test_in}
\vskip 0.15in
\begin{center}
\begin{small}
\begin{sc}
\begin{tabular}{lllllll}
\toprule
 &  & Accuracy & ECE & MCE & Train NLL & Test NLL \\
Dataset & Model &  &  &  &  &  \\
\midrule
\multirow[t]{5}{*}{CIFAR10} & V-TST, m=10 & \textbf{92.58$\pm$0.02} & \textbf{1.27$\pm$0.101} & 28.48$\pm$9.006 & 0.0671$\pm$0.0004 & 0.2617$\pm$0.0013 \\
 & V-TST, m=1 & 91.09$\pm$0.06 & 1.52$\pm$0.144 & \textbf{14.77$\pm$1.781} & 0.1125$\pm$0.001 & 0.3732$\pm$0.0043 \\
 & TST & \textbf{92.59$\pm$0.02} & 2.18$\pm$0.353 & \textbf{14.89$\pm$2.249} & \textbf{0.0573$\pm$0.0002} & \textbf{0.2482$\pm$0.004} \\
 & Temp. WRN & 92.53 & 4.4 & 26.77 & 0.063 & 0.2965 \\
 & WRN & 92.53 & 5.88 & 35.84 & 0.1038 & 0.4944 \\
\midrule
\multirow[t]{5}{*}{CIFAR100} & V-TST, m=10 & \textbf{71.93$\pm$0.04} & \textbf{5.83$\pm$0.143} & \textbf{14.03$\pm$0.59} & 0.078$\pm$0.001 & 1.202$\pm$0.0052 \\
 & V-TST, m=1 & 69.18$\pm$0.09 & 7.34$\pm$0.291 & 16.85$\pm$0.694 & 0.1094$\pm$0.0009 & 1.4315$\pm$0.0104 \\
 & TST & 71.26$\pm$0.06 & 7.07$\pm$0.084 & 16.74$\pm$1.942 & 0.0887$\pm$0.0003 & \textbf{1.1331$\pm$0.002} \\
 & Temp. WRN & 71.55 & 15.24 & 33.21 & \textbf{0.0739} & 1.3708 \\
 & WRN & 71.56 & 21.74 & 82.08 & 0.1161 & 2.2588 \\
\midrule
\multirow[t]{5}{*}{SVHN} & V-TST, m=10 & 94.76$\pm$0.01 & 0.47$\pm$0.066 & \textbf{12.67$\pm$1.958} & 0.0587$\pm$0.0003 & 0.2195$\pm$0.0012 \\
 & V-TST, m=1 & 94.73$\pm$0.02 & 0.48$\pm$0.067 & 22.63$\pm$6.889 & 0.059$\pm$0.0003 & 0.2193$\pm$0.0011 \\
 & TST & \textbf{95.1$\pm$0.01} & \textbf{0.43$\pm$0.016} & 25.04$\pm$9.617 & 0.0439$\pm$0.0001 & \textbf{0.1837$\pm$0.0004} \\
 & Temp. WRN & 95.07 & 1.88 & 19.51 & \textbf{0.0415} & 0.1926 \\
 & WRN & 95.07 & 3.48 & 26.61 & 0.0605 & 0.2869 \\
\bottomrule
\end{tabular}

\end{sc}
\end{small}
\end{center}
\end{table*}

We run a number of experiments to verify the benefits of TST and V-TST. As base models, we use over-parameterized image classifiers. We train WRN-28-10 \cite{WRN} on CIFAR10, SVHN, and CIFAR100 and additionally show that these results hold for ViT \cite{ViT} models trained on CIFAR10. We use these models as the results of training Stage 1, initializing new fully connected layers, and freezing the feature extraction layers (Convolutional layers for WRN and Attention layers for ViT). We then perform Stage 2 of TST and V-TST, training with CE loss and the ELBO respectively. Once again, training set is shared in Stages 1 and 2. We always train with Adam optimizer \cite{AdamOptim}. For comparison, we also use temperature scaling on the WRNs and include them in our evaluation for comparison to our method. We refer to \cref{supp:WRN_ViT} for details on the WRN/ViT training. \\

Once trained, we evaluate the models on metrics related to model fit and calibration. We show the accuracy, the expected calibration error (ECE) and the maximum calibration error (MCE) \cite{ECE_MCE} on the test set, and the negative log-likelihood (NLL) on both the training set and the test set. We evaluate ECE and MCE using 10 bins and report them in percentage. 

We also evaluate the models on datasets with \textit{distribution shifts}. For CIFAR10 and CIFAR100, we evaluate on CIFAR10-C and CIFAR100-C \cite{CIFAR10C} respectively, whilst for SVHN, we evaluate on rotated versions of the SVHN datasets. For details on these rotations we refer to \cref{supp:SVHN_Rotations}. Finally, we evaluate on \textit{out-of-distribution} (OOD) datasets. For the models trained on SVHN we evaluate on CIFAR10 and CIFAR100, for CIFAR10 we evaluate on CIFAR100 and SVHN and for CIFAR100 we evaluate on CIFAR10 and SVHN. We evaluate OOD detection using AUROC \cite{AUROC} and false-positive rate at 95\% (FPR95). For the base models (WRN and ViT), we only run one seed, whilst we run ten seeds of parameter re-initialization for Stage 2 for TST and V-TST and report the mean and standard error of the mean. 

For both TST and V-TST we re-initialize the fully-connected layers in Stage 2 with three FC layers. We do this to allow some flexibility in the classification layer, without overparametrizing. We test 6 dimensions over the final hidden layer output $\mathbf{z}$ in the range $[2, 512]$. We refer to \cref{supp:TST_VTST} for details on the models used in TST and V-TST and training details. We also note that in Stage 2 of both TST and V-TST, we only train for 40 additional epochs, but in most of our experiments, much less are required to converge. For all models trained we use early stopping based on the validation loss. \\

Regarding V-TST, we only use the $1$ sample in the MC approximation during training and $m$ samples in the testing phase. For stability reasons, we choose a diagonal covariance matrix in the variational family and upper-bound the variance to $1$. Additionally, we note that we downscale the KL loss by the dimension of the final hidden layer. We do this because the KL otherwise dominates the CE loss during training.

\subsection{Benchmark Results}

\begin{figure}[h]
\vskip 0.2in
\begin{center}
\centerline{\includegraphics[width=\columnwidth]{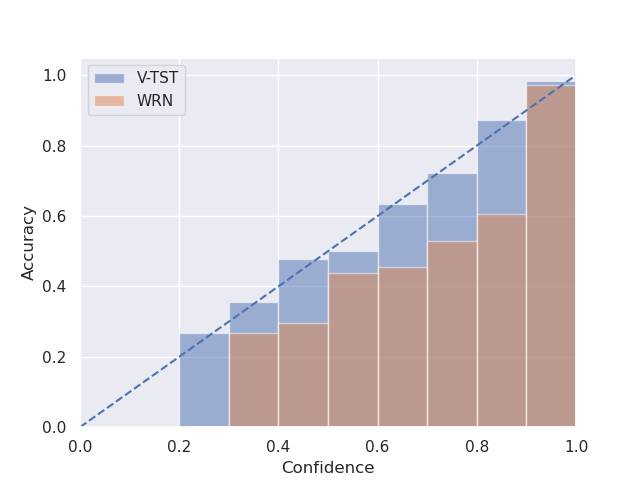}}
\caption{ECE plot of V-TST trained model and base WRN trained on SVHN.}
\label{fig:ece_plot_SVHN}
\end{center}
\vskip -0.2in
\end{figure}

\begin{figure}[h]
\vskip 0.2in
\begin{center}
\centerline{\includegraphics[width=\columnwidth]{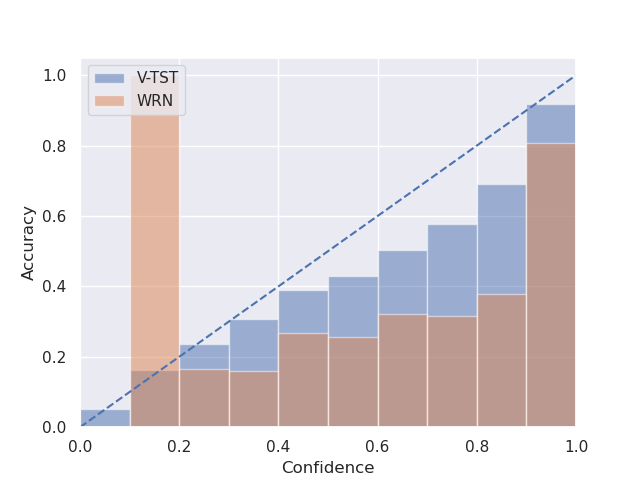}}
\caption{ECE plot of V-TST trained model and base WRN trained on CIFAR100.}
\label{fig:ece_plot_CIFAR100}
\end{center}
\vskip -0.2in
\end{figure}

In \cref{tab:best_test_in}, we show the evaluation metrics on the in-distribution data. Throughout this section, across all tables, the models listed with the same name have the same hyperparameters for a specific dataset but only show the best-performing configurations. For the full results, we refer to \cref{supp:WRN_results}. First, note that the test accuracy in all cases is comparable, except for a slight degradation in the case of V-TST when $m=1$. 
More importantly, we show that both V-TST and TST improve ECE across all datasets and, in many cases, improve the MCE. V-TST improves base WRN model ECE by $78.40\%$, $73.18\%$, and $87.64\%$ for CIFAR10, CIFAR100, and SVHN, respectively. In \cref{fig:ece_plot_SVHN} and \cref{fig:ece_plot_CIFAR100}, we compare the calibration plots for the V-TST and base WRN on  SVHN and CIFAR100 respectively to illustrate the improvements further. Here, it can be seen how V-TST improves the ECE across all bins, illustrating that improved calibration is not intrinsic to specific bins but rather across all predictions. \\

\begin{figure*}[t]
\vskip 0.2in
\begin{center}
\centerline{\includegraphics[width=\textwidth]{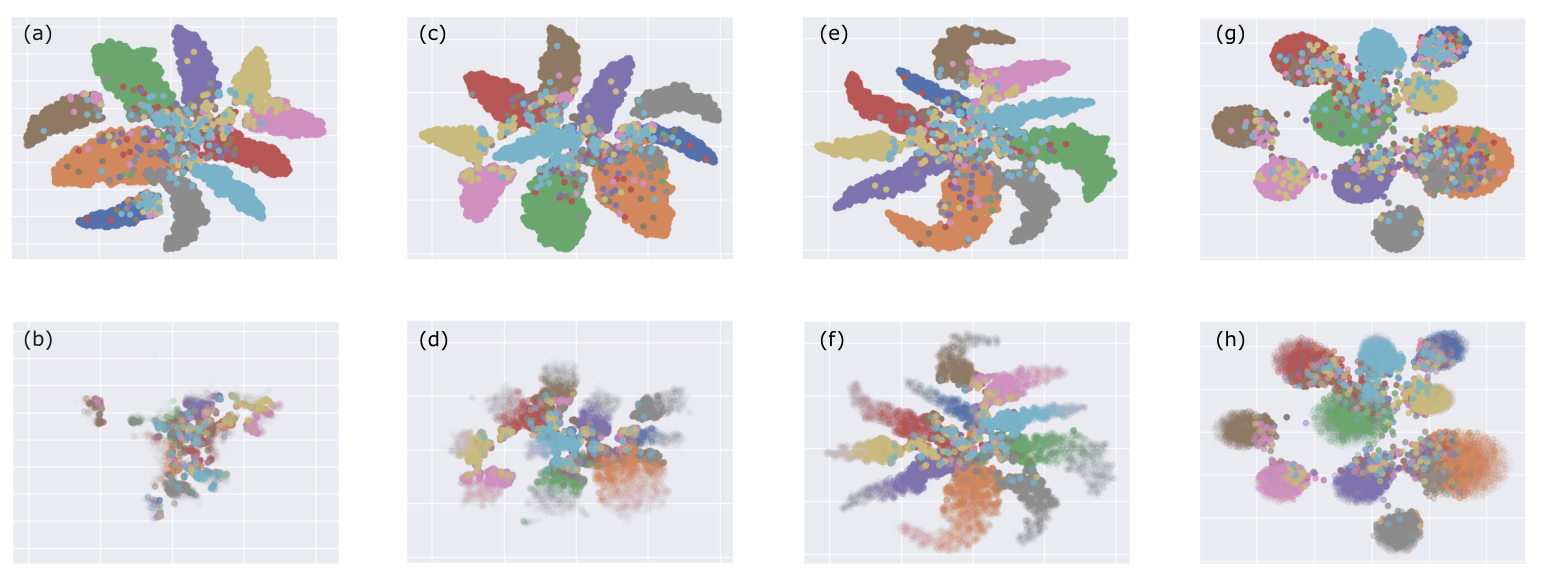}}
\caption{Last hidden layer outputs 2D t-SNE plots of test points from SVHN. (a) and (b) is from a WRN trained in \textbf{one stage}, (c) and (d) is from a \textbf{temperature scaled} version of the WRN from the first column, (e) and (f) from \textbf{TST}, and (g) and (h) is \textbf{V-TST}. In the top row we plot all test points with equal color intensity, whilst in the second row, higher color intensity indicates higher entropy of the prediction for a given point. Colors indicate which class the point truly belongs to. All these features are behind the improvements reported for TST and V-TST.}
\label{fig:latent_discussion}
\end{center}
\vskip -0.2in
\end{figure*}

Interestingly, we additionally note how both TST and V-TST decrease NLL on both the training set and the test set, which seems to indicate that training with TST is not simply a regularizer against overfitting. Had this been the case, one would expect to see a decrease in the generalization gap, i.e. an increase in the training NLL, but a decrease in the test NLL. \cref{fig:latent_discussion} illustrates the TST and V-TST effect in the structure of the hidden space $\mathbf{z}$. 
We show t-SNE plots of the WRN in (a) and (b), temperature-scaled WRN in (c) and (d), for TST in (e) and (f) and V-TST in (g) and (h) on SVHN. In the top row we plot all test points, whilst in the bottom row we again scale color intensity by $H(p(y|\mathbf{x}))$. We note how the entropy of most test points increases significantly as we go from left to right in the columns and how, in (g) and (h), we manage to get Gaussian distributions within each class.

\begin{table*}[t]
\caption{Evaluation Metrics on Shifted Data and OOD Data}
\vskip 0.15in
\begin{center}
\begin{small}
\begin{sc}
\begin{tabular}{llllll}
\toprule
 &  & SHIFT ECE & SHIFT MCE & OOD AUROC & OOD FPR95 \\
Dataset & Model &  &  &  &  \\
\midrule
\multirow[t]{5}{*}{CIFAR10} & V-TST, m=10 & \textbf{10.41$\pm$0.22} & 22.92$\pm$0.57 & 0.821$\pm$0.003 & 0.751$\pm$0.002 \\
 & V-TST, m=1 & 12.89$\pm$0.3 & \textbf{21.01$\pm$0.58} & 0.722$\pm$0.005 & 0.824$\pm$0.006 \\
 & TST & 11.62$\pm$0.75 & 25.39$\pm$1.74 & 0.874$\pm$0.002 & 0.699$\pm$0.009 \\
 & Temp. WRN & 16.45 & 36.77 & 0.872 & 0.746 \\
 & WRN & 20.27 & 45.06 & \textbf{0.891} & \textbf{0.653} \\
\midrule
\multirow[t]{5}{*}{CIFAR100} & V-TST, m=10 & \textbf{14.3$\pm$0.23} & \textbf{26.96$\pm$0.46} & 0.791$\pm$0.002 & 0.809$\pm$0.004 \\
 & V-TST, m=1 & 16.8$\pm$0.4 & 30.72$\pm$0.63 & 0.783$\pm$0.004 & 0.822$\pm$0.006 \\
 & TST & 17.44$\pm$0.08 & 28.51$\pm$0.14 & \textbf{0.823$\pm$0.002} & \textbf{0.764$\pm$0.006} \\
 & Temp. WRN & 29.58 & 47.67 & 0.778 & 0.816 \\
 & WRN & 40.47 & 60.38 & 0.785 & 0.892 \\
\midrule
\multirow[t]{5}{*}{SVHN} & V-TST, m=10 & 35.72$\pm$0.58 & 52.36$\pm$0.59 & 0.872$\pm$0.004 & 0.537$\pm$0.003 \\
 & V-TST, m=1 & 35.67$\pm$0.59 & 52.33$\pm$0.59 & 0.859$\pm$0.004 & 0.549$\pm$0.006 \\
 & TST & \textbf{30.97$\pm$0.07} & \textbf{48.39$\pm$0.09} & 0.935$\pm$0.001 & 0.415$\pm$0.006 \\
 & Temp. WRN & 37.53 & 55.09 & 0.934 & 0.461 \\
 & WRN & 47.06 & 80.65 & \textbf{0.945} & \textbf{0.344} \\
\bottomrule
\end{tabular}

\end{sc}
\end{small}
\end{center}
\label{tab:shift_metrics}
\end{table*}

In \cref{tab:shift_metrics}, we show the calibration metrics of the models when evaluated on shifted and OOD datasets. In the case of distribution shifts, we see that in all cases, TST and V-TST significantly improve upon the calibration of the WRN and outperform temperature scaling. When it comes to OOD detection tasks, 
TST and V-TST generally perform generally worse than baselines. We discuss potential reasons for this in the limitations section.

\subsection{MC Samples Effect on V-TST Performance}

\begin{figure}[h]
\begin{center}
\centerline{\includegraphics[width=\columnwidth]{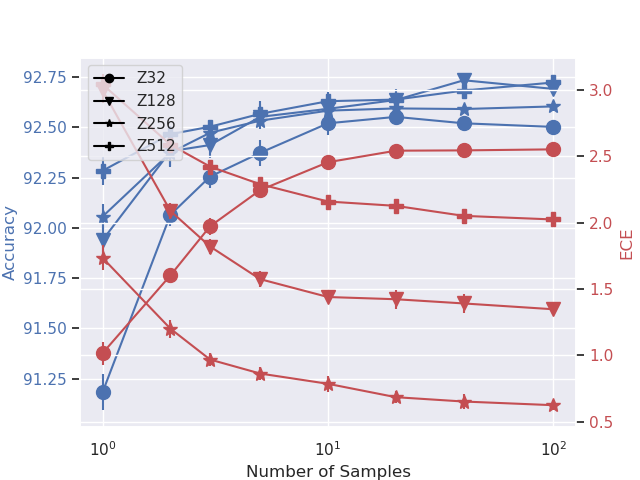}}
\caption{Effect of number of samples in Monte Carlo approximation during test evaluation of $\mathop{\mathbb{E}_{q(\mathbf{z}|\mathbf{x})}\left[p(y|\mathbf{z})\right]}$ in V-TST trained models on CIFAR10.}
\label{fig:MC_samples}
\end{center}
\vskip -0.2in
\end{figure}

Based on the previous results, a natural question that arises is how many samples are required in the MC approximation of $\mathop{\mathbb{E}_{q(\mathbf{z}|\mathbf{x})}\left[p(y|\mathbf{z})\right]}$ at V-TST prediction time. In order to verify this, we run an experiment using V-TST models trained with varying latent variable dimension on CIFAR10, and track the accuracy and ECE as we increase the number of samples used in the MC approximation. For each $m$, we run 10 seeds and report the mean and the standard error of the mean which is indicated by error bars. The results can be seen in \cref{fig:MC_samples}. Across all models, we see that as we increase the number of samples in approximation the accuracy improves. Additionally, for all models except for dimension of 32, we see that the ECE also improves as we increase the number of samples. In this regard, the results reported in \cref{tab:best_test_in} and \cref{tab:shift_metrics} can be potentially improved by tuning parameter $m$.

\subsection{Ablation Study: Under-parametrized Architecture}

To investigate our belief that TST and V-TST are beneficial when we have over-parametrized models, we run an experiment in which we define a much simpler CNN architecture than the WRN. We know that this model is not over-parametrized because it does not overfit the training data - the test and training accuracy are almost identical. We refer to \cref{supp:small_CNN} for details on this architecture. The test results are in \cref{fig:CNN_ablation}. Here, we see that the two versions of V-TST worsen the ECE, although they improve the MCE, whilst TST improves both ECE and MCE, although at the cost of accuracy. We refer to \cref{supp:CNN_results} for the full test results.

\begin{table}[h]
\caption{CNN Ablation Study Test Metrics on CIFAR10}
\label{CIFAR10}
\vskip 0.15in
\begin{center}
\begin{small}
\begin{sc}
\begin{tabular}{llll}
\toprule
 & Accuracy & ECE & MCE \\
Model &  &  &  \\
\midrule
V-TST, m=10 & \textbf{75.95$\pm$0.2} & 3.27$\pm$0.05 & \textbf{10.65$\pm$4.35} \\
V-TST, m=1 & 73.22$\pm$0.1 & 1.61$\pm$0.3 & \textbf{9.08$\pm$5.47} \\
TST & 70.77$\pm$0.2 & \textbf{0.76$\pm$0.17} & \textbf{7.79$\pm$4.67} \\
CNN & 72.48 & 1.04 & 19.89 \\
\bottomrule
\end{tabular}

\end{sc}
\end{small}
\end{center}
\label{fig:CNN_ablation}
\end{table}

\subsection{Ablation Study: Training Models End-to-End}

As part of our ablation studies, we also test whether the MLP we add at the end of the feature extraction layers and before the logits layer is the reason for the improved calibration metrics rather than TST and V-TST. Therefore, we train the best-performing versions of TST and V-TST models end-to-end on CIFAR10 and evaluate them similarly to how we have previously done. We only do this for one seed and, therefore, report no standard error of the mean. The results can be seen in \cref{fig:E2E_ablation} where we include TST and V-TST for comparison but do not report standard error of the mean. Var. E2E indicates that the adapted WRN architecture has been trained end-to-end in one step with the ELBO, while E2E is the adapted architecture trained with CE only. Here we clearly show that training in one step with the adapted architectures does not result in equivalent calibration metrics, but training variationally (Var. E2E) still improves upon E2E. This suggests that the VAE-like stochastic layer we presented for V-TST has itself regularization properties that improve model's calibration and uncertainty quantification.

\begin{table}[h]
\caption{Test Metrics for End-to-End Training on CIFAR10}
\vskip 0.15in
\begin{center}
\begin{small}
\begin{sc}
\begin{tabular}{llll}
\toprule
 & Accuracy & ECE & MCE \\
Model &  &  &  \\
\midrule
V-TST, m=10 & \textbf{92.58} & \textbf{1.27} & 28.48  \\
  V-TST, m=1 & 91.09 & 1.52 & \textbf{14.77} \\
  TST & 92.59 & 2.18 & 14.89 \\
Var. E2E, m=10 & 92.16 & 4.72 & 22.81 \\
Var. E2E, m=1 & 91.65 & 5.85 & 26.07 \\
E2E & 91.41 & 6.74 & 37.22 \\
\bottomrule
\end{tabular}

\end{sc}
\end{small}
\end{center}
\label{fig:E2E_ablation}
\end{table}

\subsection{Ablation Study: Visual Transformer}

Finally, we show that TST and V-TST does not just improve calibration in convolution architectures, but also in Visual Transformers (ViT) \cite{ViT}. We train a ViT architecture on CIFAR10, and perform Stage 2 of TST and V-TST as we did with the WRN. The results can be seen in \cref{fig:ViT_ablation}. Once again, V-TST maintains the accuracy of the ViT model, but significantly improves both the ECE and the MCE again. We refer to \cref{supp:ViT_results} for shift data and OOD data results.

\begin{table}[h]
\caption{ViT Ablation Study Test Metrics on CIFAR10}
\vskip 0.15in
\begin{center}
\begin{small}
\begin{sc}
\begin{tabular}{llll}
\toprule
 & Accuracy & ECE & MCE \\
Model &  &  &  \\
\midrule
V-TST, m=10 & \textbf{82.84$\pm$0.04} & \textbf{1.23$\pm$0.18} & \textbf{14.93$\pm$2.54} \\
V-TST, m=1 & 81.14$\pm$0.06 & 4.2$\pm$0.3 & 17.67$\pm$3.62 \\
TST & 82.3$\pm$0.06 & 1.78$\pm$0.37 & 17.71$\pm$2.2 \\
ViT & 82.56 & 13.71 & 34.27 \\
\bottomrule
\end{tabular}

\end{sc}
\end{small}
\end{center}
\label{fig:ViT_ablation}
\end{table}

\section{Limitations of TST and V-TST}

In this section we discuss concerns that may arise with TST and V-TST. Firstly, there is the computation cost of having to train in two stages. However, we find that in Stage 2, the models converge extremely quickly. In the case of TST used for the WRN architecture, the best validation loss is obtained after $1.167$ epochs on average across the three datasets. V-TST is slightly more expensive with the best validation loss for V-TST for the same datasets and models being obtained after $10.33$. The training for V-TST is also slower due to having to sample $\mathbf{z}$ in every training step, but minimally so, as we only use $m=1$ during training. At prediction time, it is necessary to increase the number of MC samples to significantly improve the performance of the V-TST models. Making predictions on the CIFAR10 test-set takes $0.932$ seconds for TST and even using $m=100$ in V-TST takes only $1.406$ seconds, a cost that may be worth it in applications where model calibration is critical. \\

An unexpected downside to TST and V-TST is that they at times reduce the performance on OOD data. To us this indicates that fitting training data well (even in terms of calibration), and training stochastically with respect to the inputs does not teach the model anything about modalities it has not seen before. An interesting future investigation would be to use BNN methods to improve OOD performance. There is much literature that lends itself readily to this for TST and V-TST, as there is much BNN research that concerns itself with only learning distributions over subsets of layer parameters, such as last layer Laplace Approximations \cite{LL_laplace}. Another future investigation could be to use the likelihood of a new test point under the prior $p(\mathbf{z})$ to perform OOD detection, although in generative models, specifically VAEs, the success of doing this has been varied \cite{VAE_outlier, VAE_outlier_2}.

\section{Conclusion}

The consensus in the DNN literature is that CE training over-parametrized models leads to poor model calibration, an effect we observe in WRN and ViT architectures. What our results indicate is that jointly training feature extraction layers and classification layers is one of the main drivers of this poor calibration. By freezing the feature extraction layers in TST and V-TST, we reduce the overall flexibility of the model. We hypothesise that this means that the classifier we train in Stage 2 cannot move data points arbitrarily far from decision boundaries to increase label likelihood artificially, as can happen when feature space and classification layers are jointly trained.

We showed that enforcing larger regularization in the feature space by adding a probabilistic prior in the $\mathbf{z}$ space significantly improved calibration even further. This is because learning a distribution over $\mathbf{z}$, $q(\mathbf{z}|\mathbf{x})$ and sampling from this at training time when predicting $y$, is an explicit augmentation of the features that is input to the logits layer. By training this way, the model learns to map several points in $\mathbf{z}$ to the same $y$, more so than it would without the variational training. Our results in the end-to-end ablation study also corroborated that variational training leads to better calibration as we saw that the V-TST architecture trained End-to-End provided increased calibration metrics in comparison to the regular end-to-end trained model. We therefore believe that this line of work of variational training for discriminative classifiers will be of interest to the community interested in calibrated classifiers.

\section{Impact Statement}

This paper presents work whose goal is to advance the field of Machine Learning. There are many potential societal consequences of our work, none which we feel must be specifically highlighted here.

\section*{Acknowledgments}
Pablo M. Olmos acknowledges the support by the Spanish government MCIN/AEI/10.13039/501100011033/FEDER, UE, under grant PID2021-123182OB-I00, and by Comunidad de Madrid under grants IND2022/TIC-23550 and ELLIS Unit Madrid.

\bibliography{TST}
\bibliographystyle{icml2024}

\newpage
\appendix
\onecolumn
\section{Training Details for WRN and ViT}
\label{supp:WRN_ViT}
For all of the datasets we train a WRN 28-10 as it is specified in \citet*{WRN}. We train it for 600 epochs using Adam optimizer with learning rate $10^{-4}$, but employ early stopping based on the validation loss. We compute the validation loss based on a validation set, which is data we split from the training set. For CIFAR10 and SVHN we use 15\% of the training set for validation, whilst for CIFAR100 we use only 5\% of the data for validation. We use data augmentation during training, but refer to the submitted code for specifics on this, and also to be able to reproduce the metrics in \cref{tab:best_test_in} and \cref{tab:shift_metrics}.\\

The ViT model is based on the implementation in \citet{ViT}. We train this model for 1000 epochs, but similarly employ early stopping based on the validation loss. Here, we also use Adam optimizer with learning rate $10^{-4}$. We use a patch size of $4$, a token dim of size $512$, depth of size $6$, $8$ heads, MLP dim of size $512$ and head dimension of size $64$. We use dropout in both the Transformer and the embeddings with $p=0.1$. This model has approximately 9.5 million parameters, where the models in \cite*{ViT} have at least 80 million parameters explaining the lower performance of our base ViT models.

\section{Training details for TST and V-TST}
\label{supp:TST_VTST}
In Stage 2 of training with TST and V-TST we re-initialize the layers parametrized by $\theta$ and $\nu$ with the same architectures for both, and we do a hyperparameter search over the dimensions of $\nu$. For the WRN architectures we re-initialize the FC layers parametrized by $\theta$ with an MLP with two layers of [input, output] size $[640, 3Z]$ and $[3Z, Z]$, where $Z$ is the size of the latent dimension, which is a hyperparameter we test varying sizes of. For all models we test $Z$ dimensions $[2, 8, 32, 128, 256, 512]$. We only use ReLU at the output of the first layer of the MLP, and do not use activation functions anywhere else. The parameters of $\nu$ is always $[Z, \text{\#classes}]$. The output of $\nu$ are logits. We only train on the same data that has been trained on in Stage 1 and still use Adam optimizer with learning rate $10^{-4}$. We still employ early stopping based on the validation loss. In the case of V-TST, we use two MLPs of the same size parametrized by $\theta$, one to output $\mu$ of the variational distribution, and one to output $\sigma$ of the variational distribution. We use a log-sigmoid activation function in the output of the MLP parametrizing $\sigma$ for stability reasons. For the ViT model, we have the same number of layers in the MLP parametrized by $\theta$, but with [input, output] sizes $[512, 3Z]$ and $[3Z, Z]$.

\section{SVHN Rotations}
\label{supp:SVHN_Rotations}
When evaluating shifted data metrics for SVHN, we evaluate on the full test-set with varying severities of rotations. We rotate with degrees $[10., 45., 90., 135., 180.]$.

\section{Small CNN Architecture}
\label{supp:small_CNN}
In the experiment in which we show that TST and V-TST do not improve under-parametrized networks, we specify a CNN with the following architecture where we use max pooling with kernel (2,2) and relu activations after all layers except the last.

\begin{itemize}
    \item Conv2D(3, 6, 5)
    \item Conv2D(6, 16, 5)
    \item Linear(400, 120)
    \item Linear(120, 84)
    \item Linear(84, 10)
\end{itemize}

\section{Additional WRN Experiment Results} 
\label{supp:WRN_results}
In all of the tables in this section, $Z$ followed by a number in the model column indicates the dimensions of the final hidden layer output. For TST and V-TST, we report the mean and standard-error of the mean over 10 seeds. $m$ is the number of samples used to approximate $\mathbb{E}_{q(\mathbf{z}|\mathbf{x})}[p(y|\mathbf{z})]$. E-to-E indicates the TST architecture trained in one stage end-to-end and Var. E-to-E indicates the V-TST model trained end-to-end in one stage.

\subsection{CIFAR10 WRN Two-Stage Results}
\newpage
\begin{table}[t]
\caption{CIFAR10 WRN In Distribution Test Results}
\vskip 0.15in
\begin{center}
\begin{small}
\begin{sc}
\begin{tabular}{llllll}
\toprule
 & Accuracy & ECE & MCE & Train NLL & Test NLL \\
Model &  &  &  &  &  \\
\midrule
E-to-E Z128 & 91.29 & 6.87 & 37.14 & 0.1036 & 0.5462 \\
E-to-E Z32 & 91.41 & 6.74 & 37.22 & 0.1141 & 0.5587 \\
TST Z128 & \bfseries 92.590$\pm$0.02 & 2.18$\pm$0.353 & 14.89$\pm$2.249 & 0.0573$\pm$0.0002 & \bfseries 0.248$\pm$0.004 \\
TST Z2 & 77.64$\pm$0.85 & 7.43$\pm$0.832 & 25.1$\pm$4.514 & 0.4522$\pm$0.02 & 0.9124$\pm$0.0199 \\
TST Z256 & \bfseries 92.590$\pm$0.01 & 2.93$\pm$0.015 & 15.86$\pm$2.857 & \bfseries 0.056$\pm$0.0001 & 0.2532$\pm$0.0004 \\
TST Z32 & 92.51$\pm$0.02 & 2.51$\pm$0.221 & 22.8$\pm$6.522 & 0.0586$\pm$0.0003 & 0.2592$\pm$0.0033 \\
TST Z512 & \bfseries 92.590$\pm$0.01 & 4.04$\pm$0.015 & 24.41$\pm$1.156 & 0.0601$\pm$0.0002 & 0.283$\pm$0.0008 \\
TST Z8 & 91.79$\pm$0.09 & 2.11$\pm$0.168 & 17.1$\pm$2.025 & 0.0739$\pm$0.0029 & 0.3083$\pm$0.0097 \\
Temp. WRN 28-10 & 92.53 & 4.4 & 26.77 & 0.063 & 0.2965 \\
V-TST Z128, m=1 & 91.9$\pm$0.04 & 2.82$\pm$0.104 & 22.27$\pm$6.658 & 0.081$\pm$0.0004 & 0.3262$\pm$0.0022 \\
V-TST Z128, m=10 & 92.58$\pm$0.02 & \bfseries 1.270$\pm$0.101 & 28.48$\pm$9.006 & 0.0671$\pm$0.0004 & 0.2617$\pm$0.0013 \\
V-TST Z2, m=1 & 33.76$\pm$0.77 & 5.83$\pm$0.517 & 20.98$\pm$1.475 & 1.6683$\pm$0.0198 & 1.7602$\pm$0.0176 \\
V-TST Z2, m=10 & 46.23$\pm$1.94 & 23.12$\pm$1.725 & 54.79$\pm$0.586 & 1.5823$\pm$0.0206 & 1.6579$\pm$0.0191 \\
V-TST Z256, m=1 & 92.14$\pm$0.05 & 2.91$\pm$0.18 & 16.04$\pm$1.074 & 0.0731$\pm$0.0007 & 0.305$\pm$0.0043 \\
V-TST Z256, m=10 & 92.56$\pm$0.03 & 1.78$\pm$0.161 & 28.01$\pm$9.02 & 0.0628$\pm$0.0003 & 0.2583$\pm$0.0025 \\
V-TST Z32, m=1 & 91.09$\pm$0.06 & 1.52$\pm$0.144 & 14.77$\pm$1.781 & 0.1125$\pm$0.001 & 0.3732$\pm$0.0043 \\
V-TST Z32, m=10 & 92.52$\pm$0.02 & 1.68$\pm$0.125 & \bfseries 13.990$\pm$1.515 & 0.0885$\pm$0.0009 & 0.2802$\pm$0.0011 \\
V-TST Z512, m=1 & 92.23$\pm$0.04 & 3.15$\pm$0.157 & 16.5$\pm$1.149 & 0.0676$\pm$0.0004 & 0.2934$\pm$0.0041 \\
V-TST Z512, m=10 & \bfseries 92.590$\pm$0.03 & 2.26$\pm$0.174 & 21.6$\pm$6.763 & 0.0601$\pm$0.0002 & 0.2572$\pm$0.0021 \\
V-TST Z8, m=1 & 88.34$\pm$0.1 & 7.09$\pm$0.187 & 19.48$\pm$2.124 & 0.2417$\pm$0.0022 & 0.4746$\pm$0.0032 \\
V-TST Z8, m=10 & 92.21$\pm$0.04 & 12.88$\pm$0.213 & 23.3$\pm$0.677 & 0.2057$\pm$0.0022 & 0.3869$\pm$0.003 \\
Var. E-to-E Z128, m=1 & 92.0 & 6.33 & 35.97 & 0.1283 & 0.6234 \\
Var. E-to-E Z128, m=10 & 92.23 & 5.66 & 37.41 & 0.097 & 0.4672 \\
Var. E-to-E Z32, m=1 & 91.65 & 5.85 & 26.07 & 0.1482 & 0.6622 \\
Var. E-to-E Z32, m=10 & 92.16 & 4.72 & 22.81 & 0.1075 & 0.4779 \\
WRN 28-10 & 92.53 & 5.88 & 35.84 & 0.1038 & 0.4944 \\
\bottomrule
\end{tabular}

\end{sc}
\end{small}
\end{center}
\vskip -0.1in
\end{table}

\begin{table}[H]
\caption{CIFAR10 WRN Shifted Data Test Results}
\vskip 0.15in
\begin{center}
\begin{small}
\begin{sc}
\begin{tabular}{lll}
\toprule
 & SHIFT ECE & SHIFT MCE \\
Model &  &  \\
\midrule
E-to-E Z128 & 20.81 & 45.58 \\
E-to-E Z32 & 20.67 & 47.13 \\
TST Z128 & 11.62$\pm$0.75 & 25.39$\pm$1.74 \\
TST Z2 & 11.2$\pm$0.58 & 22.92$\pm$2.07 \\
TST Z256 & 13.13$\pm$0.03 & 28.88$\pm$0.08 \\
TST Z32 & 12.21$\pm$0.48 & 26.43$\pm$1.08 \\
TST Z512 & 15.69$\pm$0.03 & 35.83$\pm$1.25 \\
TST Z8 & 11.66$\pm$0.31 & 23.82$\pm$0.78 \\
Temp. WRN 28-10 & 16.45 & 36.77 \\
V-TST Z128, m=1 & 13.77$\pm$0.22 & 27.28$\pm$0.49 \\
V-TST Z128, m=10 & 10.41$\pm$0.22 & 22.92$\pm$0.57 \\
V-TST Z2, m=1 & \bfseries 1.610$\pm$0.18 & 13.2$\pm$1.87 \\
V-TST Z2, m=10 & 16.44$\pm$1.31 & 52.7$\pm$1.0 \\
V-TST Z256, m=1 & 13.42$\pm$0.43 & 28.02$\pm$0.93 \\
V-TST Z256, m=10 & 10.76$\pm$0.4 & 23.97$\pm$0.94 \\
V-TST Z32, m=1 & 12.89$\pm$0.3 & 21.01$\pm$0.58 \\
V-TST Z32, m=10 & 7.33$\pm$0.24 & 16.36$\pm$0.44 \\
V-TST Z512, m=1 & 13.58$\pm$0.42 & 29.46$\pm$0.9 \\
V-TST Z512, m=10 & 11.6$\pm$0.42 & 26.03$\pm$0.95 \\
V-TST Z8, m=1 & 2.54$\pm$0.25 & \bfseries 3.810$\pm$0.25 \\
V-TST Z8, m=10 & 6.51$\pm$0.29 & 9.64$\pm$0.37 \\
Var. E-to-E Z128, m=1 & 21.73 & 41.69 \\
Var. E-to-E Z128, m=10 & 20.54 & 44.35 \\
Var. E-to-E Z32, m=1 & 21.63 & 34.72 \\
Var. E-to-E Z32, m=10 & 19.83 & 37.32 \\
WRN 28-10 & 20.27 & 45.06 \\
\bottomrule
\end{tabular}

\end{sc}
\end{small}
\end{center}
\vskip -0.1in
\end{table}

\begin{table}[H]
\caption{CIFAR10 WRN OOD Data Test Results}
\vskip 0.15in
\begin{center}
\begin{small}
\begin{sc}
\begin{tabular}{lll}
\toprule
 & OOD AUROC & OOD FPR95 \\
Model &  &  \\
\midrule
E-to-E Z128 & 0.915 & 0.561 \\
E-to-E Z32 & \bfseries 0.922 & \bfseries 0.494 \\
TST Z128 & 0.874$\pm$0.002 & 0.699$\pm$0.009 \\
TST Z2 & 0.739$\pm$0.033 & 0.856$\pm$0.036 \\
TST Z256 & 0.877$\pm$0.002 & 0.696$\pm$0.009 \\
TST Z32 & 0.879$\pm$0.003 & 0.68$\pm$0.012 \\
TST Z512 & 0.879$\pm$0.001 & 0.683$\pm$0.006 \\
TST Z8 & 0.875$\pm$0.006 & 0.68$\pm$0.026 \\
Temp. WRN 28-10 & 0.872 & 0.746 \\
V-TST Z128, m=1 & 0.775$\pm$0.004 & 0.783$\pm$0.003 \\
V-TST Z128, m=10 & 0.821$\pm$0.003 & 0.751$\pm$0.002 \\
V-TST Z2, m=1 & 0.593$\pm$0.008 & 0.924$\pm$0.003 \\
V-TST Z2, m=10 & 0.681$\pm$0.022 & 0.868$\pm$0.017 \\
V-TST Z256, m=1 & 0.802$\pm$0.007 & 0.776$\pm$0.005 \\
V-TST Z256, m=10 & 0.835$\pm$0.005 & 0.75$\pm$0.002 \\
V-TST Z32, m=1 & 0.722$\pm$0.005 & 0.824$\pm$0.006 \\
V-TST Z32, m=10 & 0.79$\pm$0.002 & 0.758$\pm$0.003 \\
V-TST Z512, m=1 & 0.823$\pm$0.005 & 0.751$\pm$0.005 \\
V-TST Z512, m=10 & 0.846$\pm$0.003 & 0.743$\pm$0.002 \\
V-TST Z8, m=1 & 0.729$\pm$0.01 & 0.843$\pm$0.011 \\
V-TST Z8, m=10 & 0.818$\pm$0.003 & 0.763$\pm$0.009 \\
Var. E-to-E Z128, m=1 & 0.772 & 0.732 \\
Var. E-to-E Z128, m=10 & 0.752 & 0.715 \\
Var. E-to-E Z32, m=1 & 0.713 & 0.809 \\
Var. E-to-E Z32, m=10 & 0.694 & 0.756 \\
WRN 28-10 & 0.891 & 0.653 \\
\bottomrule
\end{tabular}

\end{sc}
\end{small}
\end{center}
\vskip -0.1in
\end{table}

\subsection{CIFAR100 WRN Two-Stage Results}

\begin{table}[H]
\caption{CIFAR100 WRN In Dist. Results.}
\vskip 0.15in
\begin{center}
\begin{small}
\begin{sc}
\begin{tabular}{llllll}
\toprule
 & Accuracy & ECE & MCE & Train NLL & Test NLL \\
Model &  &  &  &  &  \\
\midrule
TST Z128 & 70.19$\pm$0.08 & 9.41$\pm$0.356 & 21.14$\pm$0.994 & 0.0976$\pm$0.0015 & 1.2556$\pm$0.0066 \\
TST Z2 & 6.96$\pm$0.17 & 3.5$\pm$0.197 & 47.52$\pm$1.227 & 3.3507$\pm$0.0232 & 4.0506$\pm$0.0171 \\
TST Z256 & 70.9$\pm$0.06 & 8.88$\pm$0.054 & 19.14$\pm$0.655 & 0.0875$\pm$0.0002 & 1.1923$\pm$0.002 \\
TST Z32 & 65.57$\pm$0.72 & 11.97$\pm$1.359 & 23.86$\pm$2.658 & 0.1684$\pm$0.0043 & 1.5957$\pm$0.0398 \\
TST Z512 & 71.26$\pm$0.06 & 7.07$\pm$0.084 & 16.74$\pm$1.942 & 0.0887$\pm$0.0003 & \bfseries 1.133$\pm$0.002 \\
TST Z8 & 49.52$\pm$0.35 & 9.34$\pm$0.857 & 16.83$\pm$2.705 & 0.882$\pm$0.0235 & 2.6482$\pm$0.0225 \\
Temp. WRN 28-10 & 71.55 & 15.24 & 33.21 & \bfseries 0.074 & 1.3708 \\
V-TST Z128, m=1 & 69.18$\pm$0.09 & 7.34$\pm$0.291 & 16.85$\pm$0.694 & 0.1094$\pm$0.0009 & 1.4315$\pm$0.0104 \\
V-TST Z128, m=10 & 71.64$\pm$0.07 & 3.84$\pm$0.082 & \bfseries 7.240$\pm$0.305 & 0.0945$\pm$0.0013 & 1.2129$\pm$0.0021 \\
V-TST Z2, m=1 & 2.83$\pm$0.07 & \bfseries 0.180$\pm$0.027 & 22.81$\pm$6.562 & 4.2878$\pm$0.0131 & 4.431$\pm$0.009 \\
V-TST Z2, m=10 & 4.55$\pm$0.15 & 2.52$\pm$0.141 & 33.87$\pm$13.028 & 4.2054$\pm$0.0134 & 4.3332$\pm$0.0094 \\
V-TST Z256, m=1 & 69.89$\pm$0.08 & 8.48$\pm$0.367 & 19.76$\pm$0.974 & 0.0942$\pm$0.001 & 1.3975$\pm$0.0127 \\
V-TST Z256, m=10 & 71.9$\pm$0.05 & 4.87$\pm$0.118 & 11.68$\pm$0.585 & 0.0827$\pm$0.0012 & 1.2072$\pm$0.0034 \\
V-TST Z32, m=1 & 65.79$\pm$0.09 & 3.35$\pm$0.124 & 8.61$\pm$0.466 & 0.205$\pm$0.0021 & 1.5907$\pm$0.0027 \\
V-TST Z32, m=10 & 69.82$\pm$0.06 & 4.9$\pm$0.13 & 12.56$\pm$0.454 & 0.1761$\pm$0.0019 & 1.3437$\pm$0.0026 \\
V-TST Z512, m=1 & 70.59$\pm$0.06 & 8.81$\pm$0.333 & 20.54$\pm$0.828 & 0.0858$\pm$0.0007 & 1.3416$\pm$0.0135 \\
V-TST Z512, m=10 & \bfseries 71.930$\pm$0.04 & 5.83$\pm$0.143 & 14.03$\pm$0.59 & 0.078$\pm$0.001 & 1.202$\pm$0.0052 \\
V-TST Z8, m=1 & 45.06$\pm$0.26 & 19.75$\pm$0.15 & 43.42$\pm$0.383 & 1.5655$\pm$0.0137 & 2.6529$\pm$0.0126 \\
V-TST Z8, m=10 & 53.44$\pm$0.25 & 30.45$\pm$0.135 & 55.65$\pm$0.239 & 1.4756$\pm$0.0135 & 2.4792$\pm$0.0119 \\
WRN 28-10 & 71.56 & 21.74 & 82.08 & 0.1161 & 2.2588 \\
\bottomrule
\end{tabular}

\end{sc}
\end{small}
\end{center}
\vskip -0.1in
\end{table}

\begin{table}[H]
\caption{CIFAR100 WRN Shift Results.}
\vskip 0.15in
\begin{center}
\begin{small}
\begin{sc}
\begin{tabular}{lll}
\toprule
 & SHIFT ECE & SHIFT MCE \\
Model &  &  \\
\midrule
TST Z128 & 21.13$\pm$0.52 & 33.52$\pm$0.87 \\
TST Z2 & 3.39$\pm$0.16 & 40.69$\pm$1.85 \\
TST Z256 & 20.43$\pm$0.06 & 32.76$\pm$0.1 \\
TST Z32 & 21.38$\pm$1.29 & 33.53$\pm$1.36 \\
TST Z512 & 17.44$\pm$0.08 & 28.51$\pm$0.14 \\
TST Z8 & 21.95$\pm$1.0 & 30.06$\pm$1.54 \\
Temp. WRN 28-10 & 29.58 & 47.67 \\
V-TST Z128, m=1 & 16.8$\pm$0.4 & 30.72$\pm$0.63 \\
V-TST Z128, m=10 & 9.61$\pm$0.24 & 19.16$\pm$0.4 \\
V-TST Z2, m=1 & \bfseries 0.500$\pm$0.04 & 35.29$\pm$7.44 \\
V-TST Z2, m=10 & 1.35$\pm$0.08 & 51.41$\pm$13.12 \\
V-TST Z256, m=1 & 18.16$\pm$0.46 & 33.31$\pm$0.73 \\
V-TST Z256, m=10 & 12.09$\pm$0.25 & 23.5$\pm$0.41 \\
V-TST Z32, m=1 & 12.35$\pm$0.18 & 20.97$\pm$0.34 \\
V-TST Z32, m=10 & 3.07$\pm$0.2 & \bfseries 7.380$\pm$0.3 \\
V-TST Z512, m=1 & 18.7$\pm$0.42 & 34.25$\pm$0.76 \\
V-TST Z512, m=10 & 14.3$\pm$0.23 & 26.96$\pm$0.46 \\
V-TST Z8, m=1 & 7.52$\pm$0.1 & 34.16$\pm$0.36 \\
V-TST Z8, m=10 & 15.52$\pm$0.11 & 47.98$\pm$0.36 \\
WRN 28-10 & 40.47 & 60.38 \\
\bottomrule
\end{tabular}

\end{sc}
\end{small}
\end{center}
\vskip -0.1in
\end{table}

\

\begin{table}[H]
\caption{CIFAR100 WRN OOD Results.}
\vskip 0.15in
\begin{center}
\begin{small}
\begin{sc}
\begin{tabular}{lll}
\toprule
 & OOD AUROC & OOD FPR95 \\
Model &  &  \\
\midrule
TST Z128 & 0.806$\pm$0.006 & 0.8$\pm$0.016 \\
TST Z2 & 0.612$\pm$0.021 & 0.925$\pm$0.008 \\
TST Z256 & 0.812$\pm$0.004 & 0.791$\pm$0.008 \\
TST Z32 & 0.778$\pm$0.011 & 0.829$\pm$0.019 \\
TST Z512 & \bfseries 0.823$\pm$0.002 & \bfseries 0.764$\pm$0.006 \\
TST Z8 & 0.722$\pm$0.018 & 0.862$\pm$0.025 \\
Temp. WRN 28-10 & 0.778 & 0.816 \\
V-TST Z128, m=1 & 0.783$\pm$0.004 & 0.822$\pm$0.006 \\
V-TST Z128, m=10 & 0.792$\pm$0.002 & 0.81$\pm$0.006 \\
V-TST Z2, m=1 & 0.552$\pm$0.005 & 0.94$\pm$0.002 \\
V-TST Z2, m=10 & 0.637$\pm$0.013 & 0.913$\pm$0.006 \\
V-TST Z256, m=1 & 0.788$\pm$0.002 & 0.815$\pm$0.005 \\
V-TST Z256, m=10 & 0.793$\pm$0.002 & 0.808$\pm$0.005 \\
V-TST Z32, m=1 & 0.769$\pm$0.006 & 0.84$\pm$0.01 \\
V-TST Z32, m=10 & 0.791$\pm$0.005 & 0.803$\pm$0.01 \\
V-TST Z512, m=1 & 0.791$\pm$0.002 & 0.815$\pm$0.005 \\
V-TST Z512, m=10 & 0.791$\pm$0.002 & 0.809$\pm$0.004 \\
V-TST Z8, m=1 & 0.711$\pm$0.011 & 0.868$\pm$0.009 \\
V-TST Z8, m=10 & 0.738$\pm$0.012 & 0.853$\pm$0.01 \\
WRN 28-10 & 0.785 & 0.892 \\
\bottomrule
\end{tabular}

\end{sc}
\end{small}
\end{center}
\vskip -0.1in
\end{table}

\subsection{SVHN WRN Two-Stage Results}

\begin{table}[H]
\caption{SVHN WRN In Dist. Results.}
\vskip 0.15in
\begin{center}
\begin{small}
\begin{sc}
\begin{tabular}{llllll}
\toprule
 & Accuracy & ECE & MCE & Train NLL & Test NLL \\
Model &  &  &  &  &  \\
\midrule
TST Z128 & \bfseries 95.100$\pm$0.01 & \bfseries 0.430$\pm$0.016 & 25.04$\pm$9.617 & 0.0439$\pm$0.0001 & \bfseries 0.184$\pm$0.0004 \\
TST Z2 & 87.16$\pm$0.64 & 9.38$\pm$0.424 & 25.54$\pm$1.541 & 0.3045$\pm$0.0063 & 0.6263$\pm$0.0109 \\
TST Z256 & \bfseries 95.100$\pm$0.01 & 1.43$\pm$0.01 & \bfseries 9.180$\pm$0.429 & \bfseries 0.042$\pm$0.0 & 0.1891$\pm$0.0003 \\
TST Z32 & 95.02$\pm$0.01 & 0.82$\pm$0.046 & 26.6$\pm$6.255 & 0.0504$\pm$0.0003 & 0.1922$\pm$0.0011 \\
TST Z512 & \bfseries 95.100$\pm$0.01 & 2.26$\pm$0.009 & 15.33$\pm$0.392 & 0.0433$\pm$0.0001 & 0.2028$\pm$0.0004 \\
TST Z8 & 94.65$\pm$0.04 & 0.86$\pm$0.085 & 27.18$\pm$7.062 & 0.0592$\pm$0.0009 & 0.2234$\pm$0.003 \\
Temp. WRN 28-10 & 95.07 & 1.88 & 19.51 & \bfseries 0.042 & 0.1926 \\
V-TST Z128, m=1 & 94.27$\pm$0.02 & 0.82$\pm$0.111 & 17.66$\pm$7.441 & 0.0781$\pm$0.0008 & 0.2574$\pm$0.002 \\
V-TST Z128, m=10 & 94.19$\pm$0.02 & 0.81$\pm$0.118 & 21.33$\pm$6.973 & 0.0781$\pm$0.0008 & 0.2588$\pm$0.002 \\
V-TST Z2, m=1 & 43.87$\pm$0.42 & 7.04$\pm$0.36 & 16.88$\pm$0.875 & 1.5066$\pm$0.0079 & 1.5777$\pm$0.0081 \\
V-TST Z2, m=10 & 43.79$\pm$0.41 & 6.97$\pm$0.328 & 16.4$\pm$0.914 & 1.506$\pm$0.0075 & 1.5777$\pm$0.0082 \\
V-TST Z256, m=1 & 94.5$\pm$0.02 & 0.55$\pm$0.09 & 14.43$\pm$1.996 & 0.0667$\pm$0.0004 & 0.2345$\pm$0.0015 \\
V-TST Z256, m=10 & 94.51$\pm$0.03 & 0.52$\pm$0.076 & 10.85$\pm$2.212 & 0.0669$\pm$0.0004 & 0.235$\pm$0.0013 \\
V-TST Z32, m=1 & 92.8$\pm$0.04 & 1.58$\pm$0.11 & 23.54$\pm$6.687 & 0.1202$\pm$0.0014 & 0.3214$\pm$0.0021 \\
V-TST Z32, m=10 & 92.78$\pm$0.03 & 1.54$\pm$0.132 & 24.63$\pm$6.551 & 0.1196$\pm$0.0015 & 0.3218$\pm$0.0017 \\
V-TST Z512, m=1 & 94.73$\pm$0.02 & 0.48$\pm$0.067 & 22.63$\pm$6.889 & 0.059$\pm$0.0003 & 0.2193$\pm$0.0011 \\
V-TST Z512, m=10 & 94.76$\pm$0.01 & 0.47$\pm$0.066 & 12.67$\pm$1.958 & 0.0587$\pm$0.0003 & 0.2195$\pm$0.0012 \\
V-TST Z8, m=1 & 89.16$\pm$0.08 & 7.48$\pm$0.092 & 18.18$\pm$0.402 & 0.2531$\pm$0.0018 & 0.449$\pm$0.002 \\
V-TST Z8, m=10 & 89.11$\pm$0.09 & 7.44$\pm$0.093 & 17.38$\pm$0.327 & 0.2538$\pm$0.0016 & 0.4493$\pm$0.0024 \\
WRN 28-10 & 95.07 & 3.48 & 26.61 & 0.0605 & 0.2869 \\
\bottomrule
\end{tabular}

\end{sc}
\end{small}
\end{center}
\vskip -0.1in
\end{table}

\begin{table}[H]
\caption{SVHN WRN Shift Results.}
\vskip 0.15in
\begin{center}
\begin{small}
\begin{sc}
\begin{tabular}{lll}
\toprule
 & SHIFT ECE & SHIFT MCE \\
Model &  &  \\
\midrule
TST Z128 & 30.97$\pm$0.07 & 48.39$\pm$0.09 \\
TST Z2 & 27.77$\pm$1.03 & 37.64$\pm$1.13 \\
TST Z256 & 35.13$\pm$0.06 & 53.44$\pm$0.09 \\
TST Z32 & 28.21$\pm$0.15 & 44.2$\pm$0.24 \\
TST Z512 & 39.0$\pm$0.05 & 57.24$\pm$0.06 \\
TST Z8 & 29.18$\pm$0.26 & 45.22$\pm$0.36 \\
Temp. WRN 28-10 & 37.53 & 55.09 \\
V-TST Z128, m=1 & 37.33$\pm$0.6 & 51.5$\pm$0.65 \\
V-TST Z128, m=10 & 37.28$\pm$0.62 & 51.73$\pm$0.76 \\
V-TST Z2, m=1 & \bfseries 7.940$\pm$0.4 & 15.36$\pm$1.35 \\
V-TST Z2, m=10 & \bfseries 7.940$\pm$0.41 & \bfseries 14.430$\pm$0.39 \\
V-TST Z256, m=1 & 35.95$\pm$0.55 & 51.39$\pm$0.6 \\
V-TST Z256, m=10 & 35.93$\pm$0.54 & 51.23$\pm$0.54 \\
V-TST Z32, m=1 & 38.18$\pm$0.39 & 50.43$\pm$0.5 \\
V-TST Z32, m=10 & 38.22$\pm$0.37 & 50.39$\pm$0.45 \\
V-TST Z512, m=1 & 35.67$\pm$0.59 & 52.33$\pm$0.59 \\
V-TST Z512, m=10 & 35.72$\pm$0.58 & 52.36$\pm$0.59 \\
V-TST Z8, m=1 & 28.91$\pm$0.13 & 37.36$\pm$0.2 \\
V-TST Z8, m=10 & 28.84$\pm$0.13 & 37.39$\pm$0.19 \\
WRN 28-10 & 47.06 & 80.65 \\
\bottomrule
\end{tabular}

\end{sc}
\end{small}
\end{center}
\vskip -0.1in
\end{table}

\begin{table}[H]
\caption{SVHN WRN OOD Results.}
\vskip 0.15in
\begin{center}
\begin{small}
\begin{sc}
\begin{tabular}{lll}
\toprule
 & OOD AUROC & OOD FPR95 \\
Model &  &  \\
\midrule
TST Z128 & 0.935$\pm$0.001 & 0.415$\pm$0.006 \\
TST Z2 & 0.8$\pm$0.024 & 0.757$\pm$0.031 \\
TST Z256 & 0.936$\pm$0.0 & 0.412$\pm$0.003 \\
TST Z32 & 0.933$\pm$0.002 & 0.432$\pm$0.011 \\
TST Z512 & 0.937$\pm$0.0 & 0.404$\pm$0.002 \\
TST Z8 & 0.918$\pm$0.003 & 0.495$\pm$0.018 \\
Temp. WRN 28-10 & 0.934 & 0.461 \\
V-TST Z128, m=1 & 0.816$\pm$0.004 & 0.645$\pm$0.004 \\
V-TST Z128, m=10 & 0.831$\pm$0.003 & 0.614$\pm$0.004 \\
V-TST Z2, m=1 & 0.591$\pm$0.004 & 0.926$\pm$0.001 \\
V-TST Z2, m=10 & 0.577$\pm$0.005 & 0.928$\pm$0.001 \\
V-TST Z256, m=1 & 0.844$\pm$0.004 & 0.593$\pm$0.005 \\
V-TST Z256, m=10 & 0.856$\pm$0.003 & 0.571$\pm$0.004 \\
V-TST Z32, m=1 & 0.776$\pm$0.003 & 0.741$\pm$0.004 \\
V-TST Z32, m=10 & 0.787$\pm$0.002 & 0.728$\pm$0.002 \\
V-TST Z512, m=1 & 0.859$\pm$0.004 & 0.549$\pm$0.006 \\
V-TST Z512, m=10 & 0.872$\pm$0.004 & 0.537$\pm$0.003 \\
V-TST Z8, m=1 & 0.749$\pm$0.002 & 0.8$\pm$0.002 \\
V-TST Z8, m=10 & 0.756$\pm$0.001 & 0.798$\pm$0.002 \\
WRN 28-10 & \bfseries 0.945 & \bfseries 0.344 \\
\bottomrule
\end{tabular}

\end{sc}
\end{small}
\end{center}
\vskip -0.1in
\end{table}

\section{CIFAR10 CNN Two-Stage Results}
\label{supp:CNN_results}
\begin{table}[H]
\caption{CIFAR10 CNN In Dist. Results.}
\vskip 0.15in
\begin{center}
\begin{small}
\begin{sc}
\begin{tabular}{llllll}
\toprule
 & Accuracy & ECE & MCE & Train NLL & Test NLL \\
Model &  &  &  &  &  \\
\midrule
CNN & 72.48$\pm$0.0 & 1.04$\pm$0.0 & 19.89$\pm$0.0 & 0.7132$\pm$0.0 & 0.81$\pm$0.0 \\
TST Z128 & 73.97$\pm$0.2 & 2.92$\pm$0.332 & 18.66$\pm$0.39 & 0.5951$\pm$0.0028 & 0.7593$\pm$0.0027 \\
TST Z256 & 74.86$\pm$0.11 & 3.9$\pm$0.085 & 15.29$\pm$3.719 & 0.5326$\pm$0.0063 & 0.7478$\pm$0.0007 \\
TST Z32 & 70.77$\pm$0.18 & \bfseries 0.760$\pm$0.172 & 7.79$\pm$4.67 & 0.7706$\pm$0.0025 & 0.834$\pm$0.002 \\
TST Z512 & 74.84$\pm$0.16 & 2.58$\pm$0.413 & 22.87$\pm$4.241 & 0.5596$\pm$0.0138 & 0.7397$\pm$0.0018 \\
TST Z8 & 65.51$\pm$0.15 & 2.39$\pm$0.194 & 14.45$\pm$5.075 & 0.9707$\pm$0.0044 & 0.9889$\pm$0.0055 \\
V-TST Z128, m=1 & 73.22$\pm$0.12 & 1.61$\pm$0.299 & 9.08$\pm$5.473 & 0.6416$\pm$0.002 & 0.7847$\pm$0.0022 \\
V-TST Z128, m=10 & 74.44$\pm$0.13 & 1.26$\pm$0.28 & 14.54$\pm$2.986 & 0.6111$\pm$0.0017 & 0.7474$\pm$0.0004 \\
V-TST Z256, m=1 & 74.23$\pm$0.14 & 3.65$\pm$0.278 & 22.5$\pm$4.379 & 0.5368$\pm$0.0029 & 0.7563$\pm$0.0042 \\
V-TST Z256, m=10 & 75.27$\pm$0.11 & 1.72$\pm$0.241 & 9.96$\pm$4.611 & 0.5127$\pm$0.0043 & 0.7245$\pm$0.0024 \\
V-TST Z32, m=1 & 66.74$\pm$0.28 & 1.09$\pm$0.209 & 7.81$\pm$2.272 & 0.8981$\pm$0.0069 & 0.953$\pm$0.0024 \\
V-TST Z32, m=10 & 69.9$\pm$0.13 & 6.34$\pm$0.193 & 10.59$\pm$0.682 & 0.8332$\pm$0.0052 & 0.8779$\pm$0.0027 \\
V-TST Z512, m=1 & 74.87$\pm$0.15 & 5.77$\pm$0.248 & 21.0$\pm$5.511 & 0.4442$\pm$0.0031 & 0.7735$\pm$0.0036 \\
V-TST Z512, m=10 & \bfseries 75.950$\pm$0.15 & 3.27$\pm$0.052 & 10.65$\pm$4.346 & \bfseries 0.415$\pm$0.0026 & \bfseries 0.722$\pm$0.0012 \\
V-TST Z8, m=1 & 53.54$\pm$0.53 & 3.13$\pm$0.393 & \bfseries 6.620$\pm$0.857 & 1.2905$\pm$0.0087 & 1.3107$\pm$0.0058 \\
V-TST Z8, m=10 & 61.54$\pm$0.42 & 15.53$\pm$0.429 & 22.04$\pm$0.728 & 1.1659$\pm$0.0071 & 1.1765$\pm$0.0075 \\
\bottomrule
\end{tabular}

\end{sc}
\end{small}
\end{center}
\vskip -0.1in
\end{table}

\begin{table}[H]
\caption{CIFAR10 CNN Shift Results.}
\vskip 0.15in
\begin{center}
\begin{small}
\begin{sc}
\begin{tabular}{lll}
\toprule
 & SHIFT ECE & SHIFT MCE \\
Model &  &  \\
\midrule
CNN & 9.13$\pm$0.0 & 11.8$\pm$0.0 \\
TST Z128 & 9.85$\pm$0.18 & 13.68$\pm$0.3 \\
TST Z256 & 11.03$\pm$0.26 & 15.46$\pm$0.22 \\
TST Z32 & 7.02$\pm$0.08 & 9.39$\pm$0.35 \\
TST Z512 & 9.54$\pm$0.61 & 13.37$\pm$0.83 \\
TST Z8 & 5.53$\pm$0.24 & 8.47$\pm$0.07 \\
V-TST Z128, m=1 & 8.18$\pm$0.13 & 11.15$\pm$0.08 \\
V-TST Z128, m=10 & 5.37$\pm$0.23 & 6.91$\pm$0.26 \\
V-TST Z256, m=1 & 10.23$\pm$0.34 & 14.68$\pm$0.53 \\
V-TST Z256, m=10 & 7.87$\pm$0.3 & 11.06$\pm$0.5 \\
V-TST Z32, m=1 & 6.33$\pm$0.23 & 9.0$\pm$0.31 \\
V-TST Z32, m=10 & \bfseries 0.860$\pm$0.05 & \bfseries 1.810$\pm$0.04 \\
V-TST Z512, m=1 & 13.31$\pm$0.14 & 19.75$\pm$0.25 \\
V-TST Z512, m=10 & 10.06$\pm$0.16 & 14.56$\pm$0.27 \\
V-TST Z8, m=1 & 4.23$\pm$0.04 & 5.9$\pm$0.2 \\
V-TST Z8, m=10 & 7.66$\pm$0.34 & 12.02$\pm$0.83 \\
\bottomrule
\end{tabular}

\end{sc}
\end{small}
\end{center}
\vskip -0.1in
\end{table}

\begin{table}[H]
\caption{CIFAR10 CNN OOD Results.}
\vskip 0.15in
\begin{center}
\begin{small}
\begin{sc}
\begin{tabular}{lll}
\toprule
 & OOD AUROC & OOD FPR95 \\
Model &  &  \\
\midrule
CNN & 0.704$\pm$0.034 & 0.917$\pm$0.032 \\
TST Z128 & 0.727$\pm$0.004 & 0.901$\pm$0.001 \\
TST Z256 & 0.721$\pm$0.005 & \bfseries 0.898$\pm$0.007 \\
TST Z32 & 0.73$\pm$0.003 & 0.911$\pm$0.006 \\
TST Z512 & 0.721$\pm$0.002 & 0.905$\pm$0.001 \\
TST Z8 & 0.699$\pm$0.009 & 0.934$\pm$0.008 \\
V-TST Z128, m=1 & 0.689$\pm$0.012 & 0.936$\pm$0.01 \\
V-TST Z128, m=10 & \bfseries 0.731$\pm$0.007 & 0.902$\pm$0.008 \\
V-TST Z256, m=1 & 0.685$\pm$0.011 & 0.936$\pm$0.008 \\
V-TST Z256, m=10 & 0.717$\pm$0.003 & 0.903$\pm$0.002 \\
V-TST Z32, m=1 & 0.645$\pm$0.005 & 0.946$\pm$0.005 \\
V-TST Z32, m=10 & 0.714$\pm$0.001 & 0.936$\pm$0.001 \\
V-TST Z512, m=1 & 0.688$\pm$0.003 & 0.93$\pm$0.002 \\
V-TST Z512, m=10 & 0.713$\pm$0.002 & 0.9$\pm$0.002 \\
V-TST Z8, m=1 & 0.609$\pm$0.015 & 0.944$\pm$0.006 \\
V-TST Z8, m=10 & 0.664$\pm$0.004 & 0.953$\pm$0.002 \\
\bottomrule
\end{tabular}

\end{sc}
\end{small}
\end{center}
\vskip -0.1in
\end{table}

\section{CIFAR10 ViT Two-Stage Results}
\label{supp:ViT_results}
\begin{table}[H]
\caption{CIFAR10 ViT In Dist. Results.}
\vskip 0.15in
\begin{center}
\begin{small}
\begin{sc}
\begin{tabular}{llllll}
\toprule
 & Accuracy & ECE & MCE & Train NLL & Test NLL \\
Model &  &  &  &  &  \\
\midrule
TST Z128 & 82.59$\pm$0.05 & 3.87$\pm$0.049 & 28.2$\pm$8.792 & 0.1296$\pm$0.0003 & \bfseries 0.535$\pm$0.001 \\
TST Z2 & 60.98$\pm$1.19 & 5.48$\pm$0.601 & 19.11$\pm$2.255 & 0.7726$\pm$0.0264 & 1.3832$\pm$0.0314 \\
TST Z256 & 82.84$\pm$0.04 & 7.03$\pm$0.037 & 25.24$\pm$6.103 & \bfseries 0.124$\pm$0.0002 & 0.5772$\pm$0.0009 \\
TST Z32 & 82.3$\pm$0.06 & 1.78$\pm$0.369 & 17.71$\pm$2.201 & 0.1497$\pm$0.0023 & 0.5452$\pm$0.0032 \\
TST Z512 & \bfseries 82.960$\pm$0.03 & 9.04$\pm$0.026 & 24.89$\pm$0.838 & 0.133$\pm$0.0003 & 0.6463$\pm$0.0012 \\
TST Z8 & 80.49$\pm$0.16 & 4.91$\pm$0.462 & 22.6$\pm$6.482 & 0.1756$\pm$0.0058 & 0.6622$\pm$0.008 \\
V-TST Z128, m=1 & 81.14$\pm$0.06 & 4.2$\pm$0.304 & 17.67$\pm$3.616 & 0.1616$\pm$0.0007 & 0.6538$\pm$0.0053 \\
V-TST Z128, m=10 & 82.84$\pm$0.04 & \bfseries 1.230$\pm$0.181 & 14.93$\pm$2.542 & 0.1407$\pm$0.0012 & 0.5637$\pm$0.0012 \\
V-TST Z2, m=1 & 27.05$\pm$0.4 & 2.61$\pm$0.171 & 21.94$\pm$1.933 & 1.8358$\pm$0.014 & 1.9698$\pm$0.0107 \\
V-TST Z2, m=10 & 38.64$\pm$0.91 & 18.95$\pm$0.946 & 55.29$\pm$0.73 & 1.7446$\pm$0.0152 & 1.8557$\pm$0.0132 \\
V-TST Z256, m=1 & 81.76$\pm$0.04 & 5.11$\pm$0.272 & 18.33$\pm$1.854 & 0.1486$\pm$0.0004 & 0.636$\pm$0.0065 \\
V-TST Z256, m=10 & 82.94$\pm$0.04 & 2.54$\pm$0.222 & 19.51$\pm$4.521 & 0.1319$\pm$0.0007 & 0.56$\pm$0.0026 \\
V-TST Z32, m=1 & 79.42$\pm$0.08 & 2.44$\pm$0.267 & 13.14$\pm$1.92 & 0.2116$\pm$0.0013 & 0.7357$\pm$0.0033 \\
V-TST Z32, m=10 & 82.72$\pm$0.03 & 4.06$\pm$0.274 & 10.55$\pm$0.858 & 0.1759$\pm$0.0019 & 0.5958$\pm$0.0011 \\
V-TST Z512, m=1 & 82.1$\pm$0.07 & 5.57$\pm$0.245 & 27.88$\pm$6.765 & 0.1393$\pm$0.0004 & 0.6186$\pm$0.0053 \\
V-TST Z512, m=10 & 82.92$\pm$0.04 & 3.54$\pm$0.149 & 19.28$\pm$1.854 & 0.1277$\pm$0.0006 & 0.561$\pm$0.002 \\
V-TST Z8, m=1 & 75.18$\pm$0.12 & 4.41$\pm$0.226 & \bfseries 9.430$\pm$0.367 & 0.3676$\pm$0.0022 & 0.8786$\pm$0.0036 \\
V-TST Z8, m=10 & 82.33$\pm$0.06 & 15.57$\pm$0.21 & 23.12$\pm$0.318 & 0.3108$\pm$0.0025 & 0.7181$\pm$0.002 \\
ViT & 82.56 & 13.71 & 34.27 & 0.2363 & 1.1753 \\
\bottomrule
\end{tabular}

\end{sc}
\end{small}
\end{center}
\vskip -0.1in
\end{table}

\begin{table}[H]
\caption{CIFAR10 ViT Shift Results.}
\vskip 0.15in
\begin{center}
\begin{small}
\begin{sc}
\begin{tabular}{lll}
\toprule
 & SHIFT ECE & SHIFT MCE \\
Model &  &  \\
\midrule
TST Z128 & 11.06$\pm$0.07 & 19.03$\pm$0.12 \\
TST Z2 & 8.72$\pm$0.8 & 13.68$\pm$1.06 \\
TST Z256 & 15.34$\pm$0.06 & 27.38$\pm$0.1 \\
TST Z32 & 8.16$\pm$0.48 & 14.14$\pm$0.82 \\
TST Z512 & 18.26$\pm$0.03 & 33.18$\pm$0.07 \\
TST Z8 & 12.98$\pm$0.54 & 21.08$\pm$1.02 \\
V-TST Z128, m=1 & 10.72$\pm$0.44 & 18.28$\pm$0.75 \\
V-TST Z128, m=10 & 5.44$\pm$0.35 & 10.05$\pm$0.57 \\
V-TST Z2, m=1 & 1.69$\pm$0.22 & 12.86$\pm$2.16 \\
V-TST Z2, m=10 & 13.73$\pm$0.66 & 52.9$\pm$0.85 \\
V-TST Z256, m=1 & 12.16$\pm$0.4 & 21.13$\pm$0.77 \\
V-TST Z256, m=10 & 7.87$\pm$0.31 & 13.98$\pm$0.53 \\
V-TST Z32, m=1 & 9.2$\pm$0.41 & 14.32$\pm$0.6 \\
V-TST Z32, m=10 & \bfseries 1.290$\pm$0.14 & \bfseries 3.600$\pm$0.12 \\
V-TST Z512, m=1 & 12.78$\pm$0.26 & 22.14$\pm$0.53 \\
V-TST Z512, m=10 & 9.57$\pm$0.2 & 16.68$\pm$0.43 \\
V-TST Z8, m=1 & 2.84$\pm$0.21 & 4.29$\pm$0.13 \\
V-TST Z8, m=10 & 11.66$\pm$0.2 & 15.65$\pm$0.29 \\
ViT & 25.46 & 45.2 \\
\bottomrule
\end{tabular}

\end{sc}
\end{small}
\end{center}
\vskip -0.1in
\end{table}

\begin{table}[H]
\caption{CIFAR10 ViT OOD Results.}
\vskip 0.15in
\begin{center}
\begin{small}
\begin{sc}
\begin{tabular}{lll}
\toprule
 & OOD AUROC & OOD FPR95 \\
Model &  &  \\
\midrule
TST Z128 & 0.821$\pm$0.002 & 0.757$\pm$0.006 \\
TST Z2 & 0.71$\pm$0.022 & 0.87$\pm$0.018 \\
TST Z256 & 0.816$\pm$0.001 & 0.772$\pm$0.002 \\
TST Z32 & 0.823$\pm$0.002 & 0.751$\pm$0.007 \\
TST Z512 & 0.812$\pm$0.001 & 0.791$\pm$0.003 \\
TST Z8 & 0.797$\pm$0.004 & 0.806$\pm$0.009 \\
V-TST Z128, m=1 & 0.81$\pm$0.001 & 0.769$\pm$0.002 \\
V-TST Z128, m=10 & 0.832$\pm$0.001 & 0.727$\pm$0.003 \\
V-TST Z2, m=1 & 0.573$\pm$0.006 & 0.933$\pm$0.003 \\
V-TST Z2, m=10 & 0.667$\pm$0.02 & 0.897$\pm$0.012 \\
V-TST Z256, m=1 & 0.813$\pm$0.001 & 0.765$\pm$0.002 \\
V-TST Z256, m=10 & 0.83$\pm$0.001 & 0.733$\pm$0.004 \\
V-TST Z32, m=1 & 0.789$\pm$0.001 & 0.795$\pm$0.002 \\
V-TST Z32, m=10 & 0.832$\pm$0.001 & \bfseries 0.721$\pm$0.002 \\
V-TST Z512, m=1 & 0.816$\pm$0.002 & 0.765$\pm$0.005 \\
V-TST Z512, m=10 & 0.827$\pm$0.002 & 0.744$\pm$0.006 \\
V-TST Z8, m=1 & 0.764$\pm$0.003 & 0.825$\pm$0.004 \\
V-TST Z8, m=10 & \bfseries 0.835$\pm$0.003 & \bfseries 0.721$\pm$0.011 \\
ViT & 0.797 & 0.823 \\
\bottomrule
\end{tabular}

\end{sc}
\end{small}
\end{center}
\vskip -0.1in
\end{table}

\end{document}